\documentclass[times,twocolumn,final]{elsarticle}
\usepackage{cag}

\usepackage{framed,multirow}
\biboptions{sort&compress}
\usepackage{graphicx}
\usepackage{amssymb}
\usepackage{latexsym}
\usepackage{hyperref}

\usepackage{url}
\usepackage{amsmath} 
\usepackage{xcolor}
\definecolor{newcolor}{rgb}{.8,.349,.1}
\usepackage{makecell} 
\usepackage{hyperref}

\usepackage[switch,pagewise]{lineno} %Required by command \linenumbers below

\journal{Computers \& Graphics}

\begin{document}
\nolinenumbers 
\verso{Preprint Submitted for review}

\begin{frontmatter}

\title{Semi-supervised Single-view 3D Reconstruction via Multi Shape Prior Fusion Strategy and Self-Attention}%
%\tnotetext[tnote1]{Only capitalize first
%word and proper nouns in the title.}

\author[1]{Wei \snm{Zhou}\corref{cor1}\fnref{fn1}}
\cortext[cor1]{Corresponding author;}
\emailauthor{mczhouwei12@gmail.com}{Wei Zhou}
\emailauthor{xinzheshi1@gmail.com}{Xinzhe Shi}
\emailauthor{yunfengshe1@gmail.com}{Yunfeng She}
\emailauthor{kunlongliu11@gmail.com}{Kunlong Liu}
\emailauthor{zhangyongqin@nwu.edu.cn}{Yongqin Zhang}
\author[1]{Xinzhe \snm{Shi}\fnref{fn1}}
\author[1]{Yunfeng \snm{She}}
\author[1]{Kunlong \snm{Liu}}
\author[1]{Yongqin \snm{Zhang}}
\fntext[fn1]{These authors contributed equally to this work.} 
\address[1]{School of Information Science and Technology, Northwest University, Xi'an, 710127, China}

%\received{1 February 2017}
\received{\today}
%%%% Do not use the below for submitted manuscripts
%\finalform{28 March 2017}
%\accepted{2 April 2017}
%\availableonline{15 May 2017}
%\communicated{S. Sarkar}

\nolinenumbers 
\begin{abstract}
%%%
In the domain of single-view 3D reconstruction, traditional techniques have frequently relied on expensive and time-intensive 3D annotation data. Facing the challenge of annotation acquisition, semi-supervised learning strategies offer an innovative approach to reduce the dependence on labeled data. Despite these developments, the utilization of this learning paradigm in 3D reconstruction tasks remains relatively constrained. In this research, we created an innovative semi-supervised framework for 3D reconstruction that distinctively uniquely introduces a multi shape prior fusion strategy, intending to guide the creation of more realistic object structures. Additionally, to improve the quality of shape generation, we integrated a self-attention module into the traditional decoder. In benchmark tests on the ShapeNet dataset, our method substantially outperformed existing supervised learning methods at diverse labeled ratios of 1\%, 10\%, and 20\%. Moreover, it showcased excellent performance on the real-world Pix3D dataset. Through comprehensive experiments on ShapeNet, our framework demonstrated a 3.3\% performance improvement over the baseline. Moreover, stringent ablation studies further confirmed the notable effectiveness of our approach. Our code has been released on \url{https://github.com/NWUzhouwei/SSMP}.
%%%%
\end{abstract}

\begin{keyword}
%% MSC codes here, in the form: \MSC code \sep code
%% or \MSC[2008] code \sep code (2000 is the default)
%\MSC 41A05\sep 41A10\sep 65D05\sep 65D17
%% Keywords
\KWD Single-view 3D Reconstruction\sep Semi-supervised Learning\sep Point cloud
\end{keyword}
\end{frontmatter}

\nolinenumbers 

%% main text
\section{Introduction}
\label{sec1}
As a key research challenge in computer vision, single-view 3D reconstruction aims to infer and reconstruct the corresponding 3D structure from a single 2D image. This task possesses significant application value in multiple practical domains, such as augmented reality, robot navigation, and medical imaging \cite{GERBAUD2024103947,LIU202480,wu2024reconfusion}. In recent years, deep learning techniques have spurred significant progress in single-view 3D reconstruction, with numerous data-driven methods being proposed and validated \cite{zou2024triplane,shen2024gamba}.

% \cite{GERBAUD2024103947,LIU202480,wu2024reconfusion}. In recent years, the development of deep learning techniques has spurred significant progress in single-view 3D reconstruction, with numerous data-driven methods having been proposed and validated \cite{zou2024triplane,shen2024gamba}.

\begin{figure}[htbp] % 用于单栏图像的浮动环境
    \centering
    \includegraphics[width=0.5\textwidth]{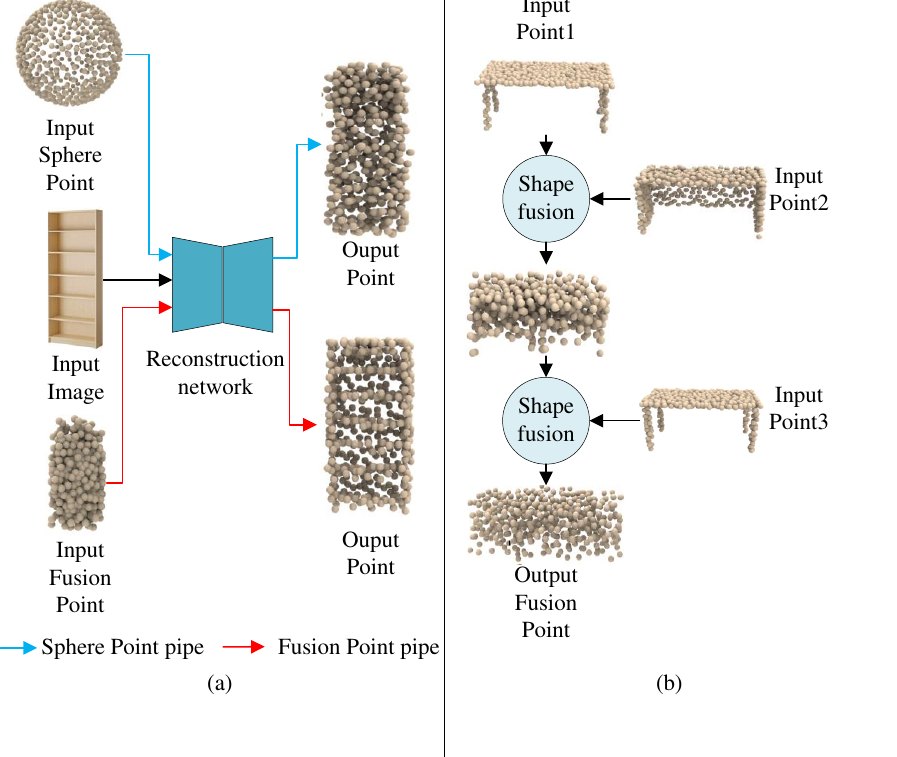}
    \caption{Comparison of reconstruction results using the initial spherical and fusion point clouds. In the initial spherical point cloud (a), there is a noticeable lack of detail, particularly in generating bookshelf layers. In contrast, the fusion point cloud (b) is derived through a multi shape prior strategy, which achieves a fusion point cloud by integrating multiple point clouds.}
    \label{fig:mean_shape}
\end{figure}
%两种方式的原理计算的体现

Despite these advances, single-view 3D reconstruction remains confronted with many challenges \cite{jin20243dfires}. Firstly, the intricate non-linear relationships between 2D images and 3D structures make model learning particularly challenging. Secondly, single-view images frequently carry limited spatial information, which leads to substantial uncertainties in the reconstruction outcomes. Furthermore, acquiring large-scale, high-quality labeled data for practical applications is typically time-consuming and expensive, imposing stringent demands on supervised learning methods \cite{szymanowicz2024splatter,melas2023pc2}.

In the fields of 2D image classification and object detection, semi-supervised learning has excelled by combining a small amount of labeled data with a vast amount of unlabeled data. This innovative approach offers new opportunities for single-view 3D reconstruction by harnessing the latent information in unlabeled data, thereby reducing the dependence on labeled data while achieving high precision. However, current semi-supervised methods for single-view 3D reconstruction rely solely on voxel-based techniques, significantly increasing GPU memory usage with higher voxel resolutions. Conversely, point-based methods only capture the surface of shapes, thus circumventing this issue. Therefore, we introduce a semi-supervised learning framework for single-view 3D point cloud reconstruction.

\begin{figure*}[htbp]  % 使用figure*环境
    \centering
    \includegraphics[width=\textwidth]{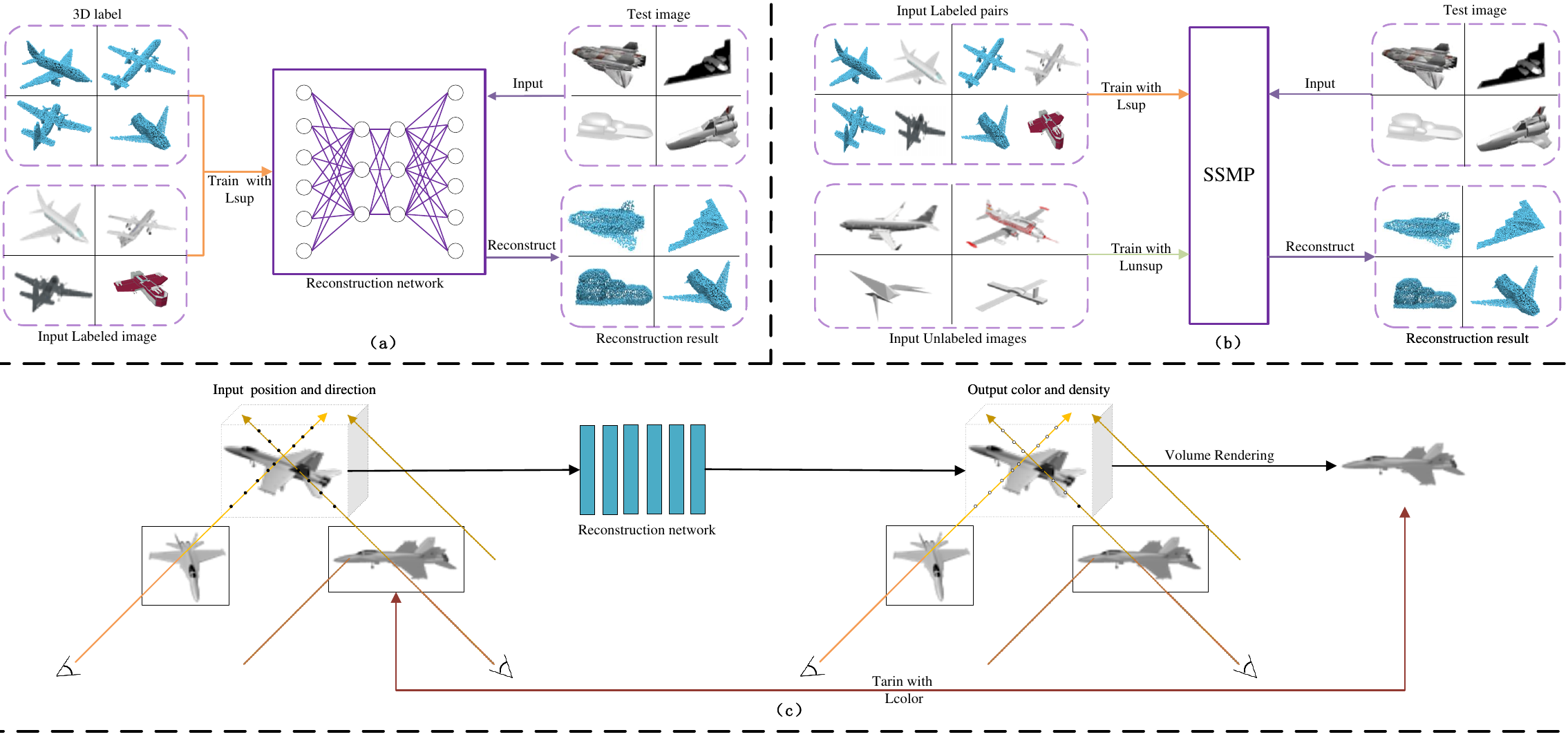}  % 插入PDF文件
    \caption{\textbf{The differences between semi-supervised learning, supervised learning and unsupervised learning.} (a) Illustration of supervised single-view 3D reconstruction, which requires a large amount of labeled data pairs.(b) Illustration of semi-supervised single-view 3D reconstruction, our proposed SSMP (Semi-Supervised Multi Shape Prior Fusion Reconstruction) model can predict the 3D shape of unlabeled images after training on a mix of a small amount of labeled and unlabeled data. (c) Unsupervised single-view 3D reconstruction requires a large amount of pose information.}
    \label{fig:example1}
\end{figure*}

As shown in \hyperref[fig:example1]{Fig. 2}, the framework leverages recent advancements in semi-supervised learning for image classification and utilizes a guided pseudo-labeling approach as its training paradigm. To produce more reliable pseudo-labels for unlabeled images and further enhance 3D reconstruction results, we address the challenge of capturing and extracting features of thinner and narrower object parts from the initial spherical point cloud, following the guidance of 3D SOC-Net~\cite{gan20233d}. As a result, we generate 3D candidate shapes using clustering algorithms (e.g., KMeans) and subsequently use the Chamfer distance as a mean shape extractor to derive a mean shape category point cloud as the initial point cloud. \hyperref[fig:mean_shape]{Fig. 1} illustrates the difference between the initial spherical point cloud (\hyperref[fig:mean_shape]{Fig. 1(a)}) and the derived mean shape category point cloud (\hyperref[fig:mean_shape]{Fig. 1(b)}). The differences are particularly noticeable in the generation of bookshelf layers when reconstructing a bookcase, where the initial spherical point cloud lacks details.

Inspired by the concept of 3DAttriFlow~\cite{wen20223d}, we utilize a deformation pipeline as the network process, reinterpreting the shape generation process as one of shape deformation. 2D image features guide the movement of each point in 3D space. We incorporate the self-attention mechanism to enhance further the decoder's capacity for processing point cloud features. This mechanism excels at capturing global dependencies among input features and augmenting their representations, all while showing exceptional robustness in dealing with noise and local anomalies.

In summary, our principal contributions are as follows:
\begin{itemize}
    \item \textbf{Semi-supervised Paradigm:} We propose a semi-supervised prior streaming 3D shape reconstruction network, the first to perform semi-supervised point cloud reconstruction using only a single image. Experiments demonstrate that by leveraging a semi-supervised learning paradigm, our model can efficiently achieve 3D point cloud reconstruction even with only a small amount of labeled data.
     \item \textbf{Multi Shape Prior Fusion Strategy:} Unlike the traditional approach of using spherical point clouds as input, we utilize a Multi Shape Prior Fusion Strategy to generate averaged point clouds. This method allows for a more comprehensive capture and fusion of features from various shapes, enabling the model to learn and recognize the corner details of 3D shapes more effectively.
    \item \textbf{Self-Attention Mechanism Decoder:} Incorporating a self-attention mechanism into the decoder enhances the model's ability to capture salient features during reconstruction. This method significantly boosts decoder performance, allowing recovery of the input 3D shape structure at a finer granularity, yielding higher-quality reconstruction results.
\end{itemize}

\section{Related Work}
\subsection{\textbf{Deep Learning for 3D Reconstruction.}}
3D reconstruction is a crucial research direction in computer vision that aims to recover the 3D structure of objects from 2D images. Methods for representing 3D structures can be categorized as either explicit or implicit. Explicit representation methods can be subdivided into voxel-based, mesh-based, and point-based approaches.

{\textbf{In voxel-based methods}, 3D-R2N2~\cite{choy20163d} uses standard 2D convolutional networks to encode input 2D images into low-dimensional embeddings, processes these embeddings with a 3D LSTM and subsequently decodes them into voxel grids via 3D convolutional networks. CIGNet~\cite{gao2023cignet} leverages category priors and intrinsic geometric relationships; by utilising reconstruction and refinement modules, it generates coarse-to-fine 3D reconstruction results from intensity images viewed from arbitrary angles. However, these voxel-based methods are constrained by the cubic growth of voxel data, which tends to result in low-resolution outputs\cite{xie2019pix2vox,xie2020pix2vox++,yagubbayli2021legoformer,tiong20223d}.

{\textbf{In mesh-based methods}}, Pixel2Mesh \cite{wang2018pixel2mesh} captures semantic information from 2D images using a mesh deformation network to transform an initial ellipsoid into the desired shape in a coarse-to-fine manner. Std-Net~\cite{mao2021std} reconstructs the structure of objects from single-view images using an autoencoder network, representing them as bounding boxes. It employs a topology-adaptive graph convolutional network (GCN) to update the vertex positions of complex topological meshes, accommodating various topological types. However, these mesh-based methods can not often represent internal or irregular structures~\cite{wen2022pixel2mesh++,yang2023single,zhang2024t}.

{\textbf{In point-based approaches}, PSGN~\cite{fan2017point} addresses the permutation invariance issue of point clouds by utilising Chamfer Distance (CD) and Earth Mover's Distance (EMD) as loss functions. Part-Wise AtlasNet~\cite{yu2022part} employs a generator-discriminator architecture, introducing two types of conditional adversarial losses to facilitate global semantic reconstruction. These point-based methods typically require lower memory usage while providing more detailed internal structural information.
However, they generally rely on using spherical point clouds as input, which presents limitations, particularly in reconstructing complex and intricate structures. Consequently, {they often perform inadequately in handling the details of complex objects~\cite{afifi2020pixel2point,melas2023pc2,wen20223d,gan20233d}.

On the other hand, recent advancements in \textbf{Neural Radiance Fields (NeRF)}~\cite{mildenhall2021nerf,hong2023lrm,lin2023vision,yu2021pixelnerf,barron2021mip,metzer2023latent} have enabled implicit representation methods to optimise neural radiance fields through differentiable rendering, minimising the discrepancy between synthesised and authentic observed images. 

Beyond NeRF, recent research \cite{poole2022dreamfusion,jun2023shap,nichol2022point}has leveraged \textbf{diffusion models} to generate three-dimensional structures from single images. For example, Zero123~\cite{liu2023zero} fine-tunes a stable diffusion model to create new views based on relative camera poses. One-2-3-45~\cite{liu2024one} builds on Zero123 by extracting 2D features from multi-view images and combining them with camera poses to construct a 3D cost volume, which a 3D convolutional neural network then processes to infer the underlying geometric structures. Wonder3D~\cite{long2024wonder3d} integrates input images, CLIP-generated text embeddings, multi-view camera parameters, and a domain switcher to produce consistent multi-view normal maps and color images. It employs a novel standard fusion algorithm to reconstruct high-quality 3D geometric structures and generate high-fidelity textured meshes from these 2D representations.

However, explicit representation methods often rely on supervised approaches that require substantial 3D data. In contrast, implicit representation methods typically use self-supervised techniques but need numerous images from different viewpoints, as well as corresponding camera poses, leading to suboptimal performance under limited viewpoint conditions.

\subsection{\textbf{Deep Learning With Semi-supervised.}}
Semi-supervised learning (SSL) 's core challenge lies in effectively utilizing labeled and unlabeled data to train algorithms. Existing methodologies can be broadly classified into two categories: entropy minimization \cite{grandvalet2004semi,lee2013pseudo,pham2021meta,zoph2020rethinking} and consistency regularization \cite{berthelot2019remixmatch,berthelot2019mixmatch,gong2021alphamatch,xie2020unsupervised,miyato2018virtual}. The entropy minimization approach is derived from self-training, which involves assigning pseudo-labels to unlabeled data and combining these pseudo-labels with manually labeled data for further training. In contrast, consistency regularization assumes that the predictions for unlabeled data should remain invariant under different perturbations. To achieve this, data augmentation is introduced, applying random perturbations to expand the training set, thereby enhancing the performance and robustness of the algorithms. Common simple augmentation techniques include random flipping, geometric transformations, and contrast adjustments of images. Additionally, more complex augmentation strategies such as Mixup, which mixes two images at random proportions, and Cutout \cite{devries2017improved}, which replaces randomly selected pixel values in an image with zeros, are also employed. 

In single-view 3D reconstruction, \cite{yang2018learning} was the first to attempt reconstructing 3D objects using a limited amount of annotated data, addressing issues of pose invariance and viewpoint consistency with a few additional camera poses. The semi-supervised soft rasterization (SSR) \cite{laradji2021ssr} employs a Siamese network, introducing extra image contours as part of the unsupervised loss. SSP3D\cite{xing2022semi} pioneered a 3D semi-supervised learning method that solely utilizes single-view images, achieving 3D voxel generation by incorporating explicit shape priors, a shape discriminator, and a prototype shape prior module. However, to our knowledge, SSL based solely on point representations has yet to be explored in the context of 3D reconstruction.

\section{Method}
As shown in \hyperref[fig:example2]{Fig. 3}, inspired by the SSP3D\cite{xing2022semi} approach, our framework SSMP consists of two training stages: the warm-up stage and the teacher-guided stage. In the warm-up stage, a limited labeled dataset $\boldsymbol{D_L} = \{(x^{l}_{i}, y^{l}_{i})\}$ is used to train the teacher model. In the teacher-guided stage, the teacher model first generates corresponding pseudo-labels for the unlabeled dataset $\boldsymbol{D_U} = \{(x^{u}_{i})\}$. Then, a student model initialized from the pre-trained teacher model is trained on both $\boldsymbol{D_L}$ and $\boldsymbol{D_U}$ to enhance performance. Meanwhile, each stage uses the fusion point cloud as input to ensure data consistency and model generalization capability.

\begin{figure*}[ht]  % 使用figure*环境
    \centering
    \includegraphics[width=\textwidth]{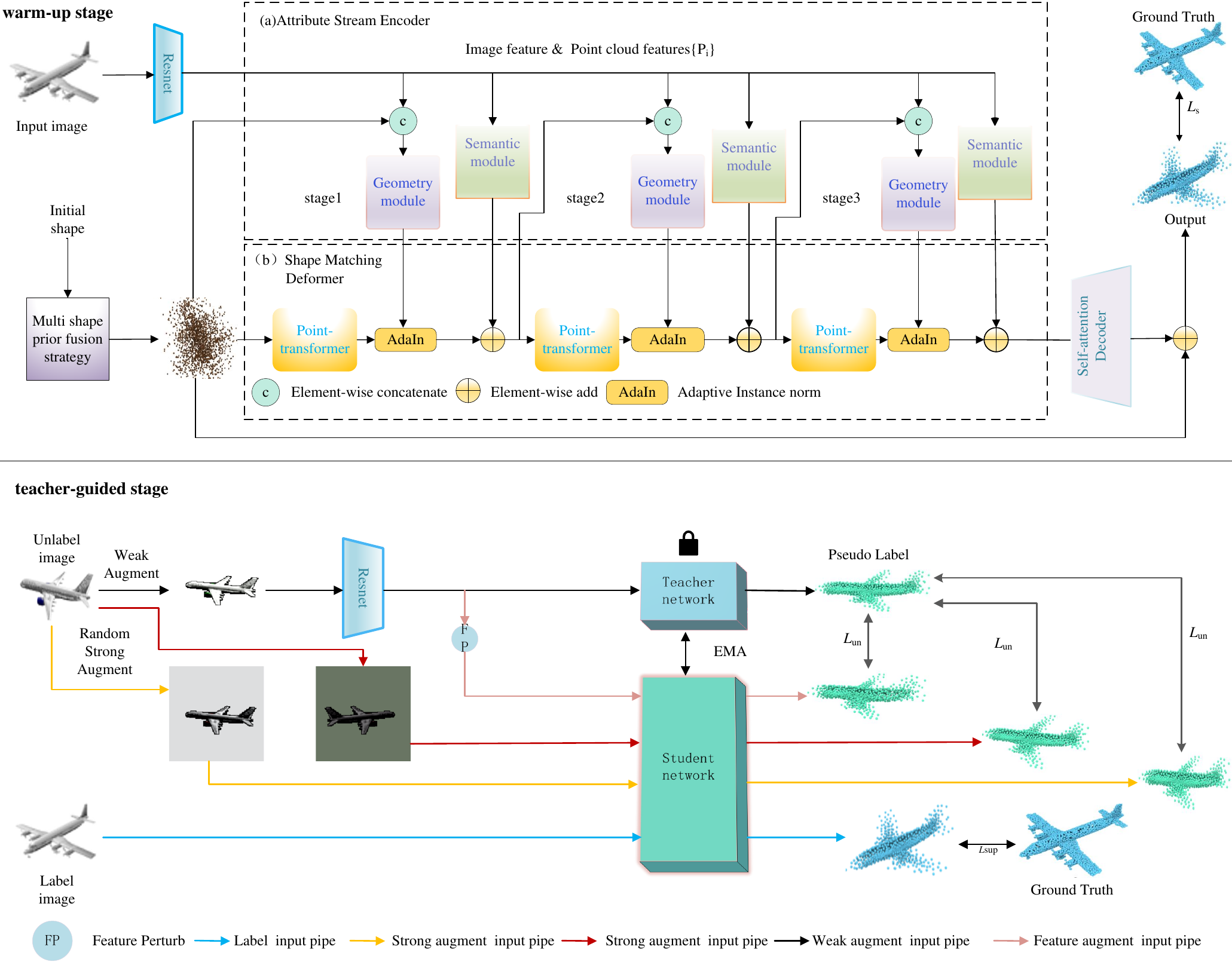}  % 插入PDF文件
    \caption{SSMP consists of two stages. The warm-up stage: we train the 3D reconstruction network using the available supervised data and the fusion shape point clouds as the initial point clouds. The teacher-guided stage: we apply three sets of image-level augmentations for the unsupervised data and one set of feature-level augmentations. With fixed parameters, the teacher generates pseudo-labels to train the student using weakly augmented data. Meanwhile, the student learns knowledge by inputting two sets of strongly augmented data and one set of feature-level augmented data. The knowledge that the student learns online is gradually transferred to the teacher's weights in a replicated mode using an exponential moving average (EMA).}
    \label{fig:example2}
\end{figure*}
Simultaneously, the teacher model employs an exponential moving average (EMA) update strategy to temporally integrate the weights of the student model, thereby generating more accurate pseudo-labels. In the following text, we first introduce the model components proposed during the warm-up stage \hyperref[sec:method1]{(Section 3.1)}, and then demonstrate how the teacher-guided stage implements the semi-supervised strategy \hyperref[sec:method2]{(Section 3.2)}. Finally, we describe the 
optimization of the model \hyperref[sec:method3]{(Section 3.3)}.

\subsection{\textbf{Warm-up Stage}}
\label{sec:method1}
As shown in \hyperref[fig:example2]{Fig. 3}, SSMP comprises four main components: an image encoder, an attribute flow encoder, a shape matching deformer, and a self-attention decoder. The image encoder utilizes a pre-trained ResNet50, while the attribute flow encoder and shape matching deformer are designed and enhanced based on 3DAttriFlow\cite{wen20223d}.
At this stage, the teacher model is trained using a standard supervised learning approach on a limited labeled dataset $\boldsymbol{D_L}$ and fusion point cloud. The following sections will introduce the modules above.
%multi shape prior fusion strategy

\textbf{Multi Shape Prior Fusion Strategy.} Single-view 3D reconstruction of point clouds frequently encounters challenges due to insufficient depth information, resulting in ambiguity during the 2D-to-3D reprojection process.  To enhance the accuracy of point cloud generation, an initial point cloud is typically introduced before feature extraction. For example, Ahmed J. et al.\cite{afifi2020pixel2point} proposed a spherical initial point cloud that uniformly distributes data points, achieving relatively accurate point cloud generation. However, spherical initial point clouds may struggle to effectively capture and extract features from thin and narrow sections of an object, potentially leading to the omission of critical components in the reconstructed 3D shapes. In research of 3D recognition, Maxim Tatarchenko et al.\cite{Tatarchenko_Richter_Ranftl_Li_Koltun_Brox_2019} found that when both the object's shape and the input average shape are well-defined, recognition methods can mitigate the risk of generating unstructured shapes. Building upon this insight, we utilize an average point cloud as input to address this issue, facilitating the reconstruction of more complete and detailed 3D shapes.

Unlike the strategy in \cite{Tatarchenko_Richter_Ranftl_Li_Koltun_Brox_2019}, which obtains shape prototypes by directly performing K-means clustering on 3D objects, our approach uses a pre-trained 3D auto-encoder to obtain shape prototypes through compact latent representations. The advantage of this approach is that the latent representations learned by the auto-encoder can capture the main features and structural information of point cloud data, enabling K-means clustering to operate within a richer feature space, thereby obtaining more representative shape prototypes.

Specifically, the encoder maps the input point cloud \( P \in \mathbb{R}^{N \times 3} \) into a low-dimensional feature space, where \( N \) is the number of points in the point cloud. After obtaining the compact latent representation of the point cloud, we apply the K-means clustering algorithm in this compact feature space, partitioning these features into \( K \) clusters. Each cluster center represents a shape prototype feature. Finally, the decoder reconstructs these cluster centers into shape prototypes \(\hat{P}_k\):
\begin{equation}
\hat{P}_k = f_{\text{dec}}\left( \text{KMeans}(f_{\text{enc}}(P), K)_k \right), \quad k = 1, 2, \ldots, K
\end{equation}
where \(\mathit{f}\) are the pretrianed 3d autoencoder. 

In [43], the average shape is derived using a straightforward averaging method applied to the obtained voxel shape prototypes, as described by the formula:
\begin{equation}
\hat{v}_{k} = \frac{1}{K} \sum_{k=0}^{K} v_{k}
\end{equation}
where \( v \) represents the voxel shape prototypes, and \(\hat{v}\) is the averaged voxel shape. However, for point-based methods, point clouds' unordered and rotation-invariant nature poses a significant challenge. Simple averaging can lead to overly sparse or dense average shapes that do not accurately represent the true shape. To tackle this issue, we incorporate the Chamfer distance as a weighting factor for the following reasons:

The Chamfer distance measures the similarity between point clouds by calculating the nearest neighbour distance for each point in one cloud to the other. The Chamfer distance increases for points with larger discrepancies, thereby reducing their influence in the weighting calculation. This ensures that points with greater distances contribute less to the average shape, avoiding overly sparse or dense results. This method preserves geometric details and accurately reflects the common structural information between different point clouds. The formula is as follows:
\begin{equation}
 P_o = \frac{\sum_{i=0}^K  P_i \cdot \frac{1}{d_{CD}(P_i, P_{\text{avg}})}}{\sum_{j=0}^K \frac{1}{d_{CD}(P_j, P_{\text{avg}})}} 
\end{equation}

\textbf{Attribute Flow Encoder.} As illustrated in Figure \hyperref[fig:example2]{3(a)}, the attribute flow encoder extracts image features and point cloud features \(P_{\mathrm{i}}\) via a geometry module and a semantic module. These features are subsequently translated into geometric information \(\{\gamma_{\mathrm{i}}, \beta_{\mathrm{i}}\}\) and semantic features \(S_{\mathrm{i}}\), where \(i\) denotes the \(i\)-th stage.

In the geometric module, image features \(x\) are concatenated with the point cloud features \(P_{\mathrm{i}}\) from stage \(i\). For the initial stage, we use \(P_{o}\) directly as the point cloud features. These combined features are then processed through a Multi-Layer Perceptron (MLP) and reshaping operations to extract geometric information \(\{\gamma_{\mathrm{i}}, \beta_{\mathrm{i}}\}\):

 \begin{equation}
\{\gamma_{\mathrm{i}}, \beta_{\mathrm{i}}\} = \text{MLP}(x : P_{\mathrm{i}})
\end{equation}
where ":" denotes feature concatenation. The resulting geometric features \(\{\gamma_{\mathrm{i}}, \beta_{\mathrm{i}}\}\) are subsequently input into the shape matching deformer, guiding the reconstruction of three-dimensional shape attributes.

In the semantic module, the extracted image features \(x\) are compressed into attribute encodings \(z_{\mathrm{i}}\) via an MLP. According to EigenGAN \cite{he2021eigengan}, for the \(j\)-th dimension activation \(z_{\mathrm{ij}}\) of the attribute encodings \(\{z_{\mathrm{i}}\}\), orthogonal basis \(u_{ij} \in \mathbb{R}^{N \times C_{i}}\) from  a linear subspace \(\mathcal{U}_{i} = \{u_{ij}\}\) are used to discover the semantic attributes behind \(z_{\mathrm{ij}}\), where \(C_{i}\) is the number of features at stage \(i\). The specific formula is as follows:
 \begin{equation}
\hat{z}_{i j}=\ell_{i j}z_{i j}u_{i j}
\end{equation}
where \(\ell_{ij}\) are learnable weights representing the importance of the semantic attributes discovered through the orthogonal bases \(u_{ij}\). By aggregating all dimensions of the attribute encodings \(z_{\mathrm{i}}\), the semantic sub-process outputs semantic features \(S_{\mathrm{i}}\) that encode explicit attribute information:
 \begin{equation}
 S_{i}\ =\sum_{j}\hat{z}_{i j}+b_{i}
\end{equation}

where \(b_{i}\) is a learnable bias. The final semantic features \(S_{i}\) are subsequently input into the shape matching deformer to guide the reconstruction of three-dimensional semantic attributes.

\textbf{Shape Matching Deformer.} The structure of the shape matching deformer is illustrated in \hyperref[fig:example2]{Fig. 3(b)}. The input to this deformer is the average point cloud \(P_{o}\), and the output is a set of displacement feature vectors \(P_i\). To more accurately predict the feature vectors \(P_i\) for each point, we employ PointTransformer instead of the graph attention module used in 3DAttriFlow \cite{wen20223d}. PointTransformer can dynamically adjust the relationships between points, thereby enhancing the flexibility and accuracy of feature representation. 
. For convenience, we denote the point features generated by Point Transformer in stage i as \(Q_i\) = \(\{q_{ik}\}\).

In the \(i\)-th stage, the shape matching deformer takes geometric information \(\{\gamma_{\mathrm{i}}, \beta_{\mathrm{i}}\}\) and semantic features \(S_{\mathrm{i}}\) as input, and adjusts the geometric information through adaptive instance normalization:

 \begin{equation}
\hat{q}_{i k}=\gamma_{i k}\cdot \frac{q_{i k}-\mu(q_{i k})}{\sigma(q_{i k})}+\beta_{i k}
\end{equation}

where \(k\) represents the \(k\)-th row vector, and \(P_{i k}\) denotes the point cloud features at stage \(i\). \(\mu(q_{i k})\) and \(\sigma(q_{i k})\) are the mean and standard deviation of \(q_{i k}\) estimated using a moving average algorithm. 

According to the geometric features, the adjusted point features are passed through the MLP layer and integrated into the point cloud features \(P_ik\) by summing them with the semantic feature \(S_i\), as follows:
 \begin{equation}
 {P}_{i k} = \hat{q}_{i k} + \phi(S_i | \theta_{S_i})
\end{equation}
where \(\phi\) represents the MLP layer, and \(\theta_{S_i}\) represents the weights of the MLP layer used to generate \({P}_{i k}\).The final point cloud features are transformed into the point cloud by passing through the self-attention decoder.

\textbf{Self-Attention Decoder.} Traditional MLP decoders have limitations in integrating global contextual information and adaptability. Previous works often avoided using self-attention mechanisms due to their high computational complexity and the potential for over-fitting, especially with large-scale point cloud data. Early attempts to integrate attention mechanisms into decoders were sometimes sub-optimal because the network architectures were not adequately designed to leverage the strengths of self-attention.

To address these limitations, our decoder employs self-attention mechanisms. This approach effectively captures global information and the relationships between features, thereby enhancing the decoder's global perception capabilities for more accurate and efficient point cloud displacement decoding. Specifically, we carefully design the self-attention layers to be computationally efficient. We use multi-head self-attention to capture diverse aspects of global context and feature relationships while applying layer normalization to enhance the model's generalization ability.

Within the self-attention decoder, the point cloud feature vectors \(P_i\) of the point cloud are first introduced into the self-attention mechanism. The core operations of this mechanism encompass the calculation of the query (\(Q\)), key (\(K\)), and value (\(V\)) vectors:
 \begin{equation}
\begin{cases}{{Q=W_{Q} \cdot P_{i}}}\\ {{K=W_{K} \cdot P_{i}}}\\ {{V=W_{V} \cdot P_{i}}}\
\end{cases}
\end{equation}
where \(W_{Q}\), \(W_{K}\), and \(W_{V}\) are the learned weight matrices. Next, we calculate the dot product similarity between the queries and keys, followed by scaling and applying the Softmax operation to obtain the attention weights:
 \begin{equation}
{\mathrm{Attention}}(Q,K,V)={\mathrm{softmax}}\left({\frac{Q \cdot K^{T}}{\sqrt{d_{k}}}}\right) \cdot V
\end{equation}
Where \(d_{k}\) is the dimension of the key vector used for scaling the dot product similarity. Finally, The point cloud displacement obtained through the attention mechanism is added to the initial point cloud \( P_o \) to generate the final reconstructed point cloud \( P \):
\begin{equation}
P=P_{o}+\mathrm{MLP}({\mathrm{Attention}}(Q,K,V))
\end{equation}

\subsection{\textbf{Teacher-guidance Stage}}
\label{sec:method2}

As shown in \hyperref[fig:example3]{Fig. 4}, unlike SSP3D, which only applies strong perturbations at the image level, we introduce feature-level perturbations to expand the perturbation space further. This approach provides more detailed and effective supervision signals than image-level perturbations, helping the student model better understand and learn the complex patterns in the data.

\begin{figure}[htbp]  % 使用figure*环境
    \centering
    \includegraphics[width=0.5\textwidth]{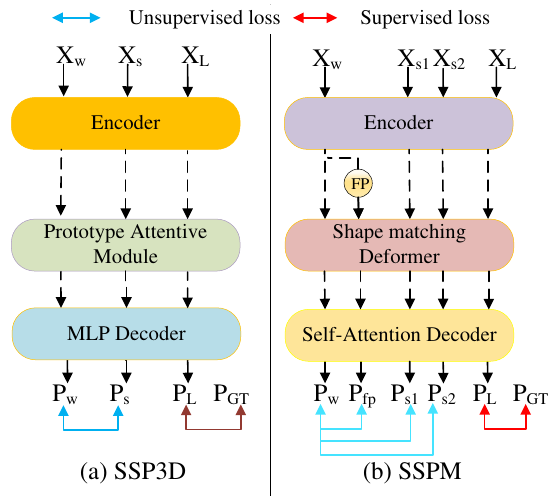}  % 插入PDF文件
    \caption{(a) SSP3D baseline. (b) Our proposed feature perturbations method (SSMP). ``FP" denotes feature perturbation, the blue line indicates the unsupervised loss, and the red line indicates the supervised loss.}
    \label{fig:example3}
\end{figure}

Once the teacher model converges in the warm-up stage, it generates pseudo-labels for the unlabelled images to supervise the student model. We initialise the student model with the weights of the teacher model and design three branches to train the student network using the unlabelled dataset \(\boldsymbol{D_U}\) and fusion point cloud. First, we perform weak data augmentation on the input images to obtain the weakly perturbed image \(X_{w}\). Then, \(X_{w}\) is fed into the image decoder to produce the weak image \(e_{w}\), while feature perturbation generates \(e_{fp}\).
\begin{equation}
\begin{cases}
{{e_{w}=g(X_{w})}}\\ {{e_{f p}=FP(e_{w})}}
\end{cases}
\end{equation}
where \(g\) is the image encoder and \(FP\) is the feature perturbation, such as dropout or adding uniform noise. 

Subsequently, the two sets of features are processed through the network to obtain the pseudo-labels \( P_{w}\) and the feature-perturbed point cloud \( P_{fp}\). On the other hand, the student’s input images undergo two different strong data augmentations to yield the strongly perturbed images \( X_{s1}\) and \( X_{s2}\). After passing through the network, we obtain the image-perturbed point clouds \( P_{s1}\) and \( P_{s2}\). The point clouds derived from these three perturbations and pseudo-labels are used to train the student network. To obtain more refined pseudo-labels, we use EMA (Exponential Moving Average) to update the teacher model based on the student’s weights throughout the teacher-guidance stage. The update rule is defined as follows:
\begin{equation}
\Psi_{tea}= \alpha\Psi_{tea}+(1-\alpha)\Psi_{stu}
\end{equation}
where \(\alpha\) is the momentum coefficient. To stabilise the training process, we gradually increase \(\alpha\) to 1 using a cosine schedule.

\subsection{\textbf{Model Optimization.}} \label{sec:method3}   
The training process is completed in two stages. In the warm-up stage, we jointly utilise reconstruction loss and regularisation loss to train the teacher model on \(D_{L}\). In the teacher-guided stage, the parameters of the teacher model are fixed, and we optimise the student network solely through supervised and unsupervised losses. In the 3D reconstruction network, the reconstruction predictions and ground truth are represented as point clouds for the reconstruction loss. We follow previous work and use Chamfer loss as the reconstruction loss function:

\begin{equation}
\begin{aligned}
\mathcal{L}_{\mathrm{CD}}(P_{r},P_{gt}) &= \frac{1}{2N}\sum_{p_{r}\in P_{r}} \operatorname*{min}_{p_{gt}\in P_{gt}} \|p_{r}-p_{gt}\|_{2} \\
& \quad + \frac{1}{2N}\sum_{p_{gt}\in P_{gt}} \operatorname*{min}_{p_{r}\in P_{r}} \|p_{gt}-p_{r}\|_{2}\label{con:cd}
\end{aligned}
\end{equation}
where \(P_{r}\) and \(P_{gt}\) represent the predicted point cloud and the ground truth point cloud, respectively, while \(N\) denotes the number of points in the point cloud. To ensure the orthogonality of \(\mathcal{U}_{i}\), we employ a regularization loss, defined as follows:
\begin{equation}
{\mathcal{L}}_{\mathrm{{Orth}}}=\sum_{i\in 1,2,3}\|{\mathcal{U}}_{i}^{\mathsf{T}}{\mathcal{U}}_{i}\|-1.
\end{equation}

The total training loss during the warm-up stage is expressed as:
\begin{equation}
\mathcal{L}_{s} = \mathcal{L}_{CD}+ \xi{\mathcal{L}}_{\mathrm{{Orth}}}
\end{equation}
where \(\xi\) is the balancing factor that determines the weight of \(\mathcal{L}_\mathrm{Orth}\), set to 100 in this paper.

In the second phase of the teacher-guided loss, we use different loss functions for supervised and unlabeled data. For supervised data, we use the loss function as shown in Eq. \ref{con:cd}. For unlabeled data, we use the following loss function:
\begin{equation}
\begin{aligned}
\mathcal{L}_{un}&=\lambda\mathcal{L}_{CD}( P_{w},P_{fp}) \\& \quad +\frac{\nu}{2}(\mathcal{L}_{CD}( P_{w},P_{s1})+\mathcal{L}_{CD}( P_{w},P_{s2})) 
\end{aligned}
\end{equation}
where \(\lambda\) and \(\nu\) are the loss weights, set to 0.5.The loss function for training the student model is shown as follows:
\begin{equation}
\mathcal{L}=\mathcal{L}_{un}+\mathcal{L}_{sup}
\end{equation}

By joint training with both supervised and unsupervised losses, we can effectively leverage both labeled and unlabeled data, resulting in improved performance.

\section{Experiment}
\subsection{\textbf{Experimental Setup.}}    
\textbf{Datasets and Evaluation Metrics.} We used the ShapeNet \cite{chang2015shapenet} and Pix3D \cite{sun2018pix3d} datasets in our experiments. For ShapeNet, we utilized 12 categories with 36,169 3D models. The training set was randomly divided into supervised and unlabeled data based on labeled sample ratios of 1\%, 10\%, and 20\%. Pix3D is a publicly available dataset that aligns real-world images with 3D models, and we followed the standard S1 split, which includes 7,539 training images and 2,530 testing images. We randomly selected 10\% of the training set as labeled data, with the remainder used as unlabeled data. The evaluation metric is the L1 chamfer distance as defined in Eq. (\ref{con:cd}).

\textbf{Implementation Details.} The experiments in this paper were conducted in Python 3.7.9, utilizing PyTorch 1.11.0 and CUDA 11.7. The operating system used was Ubuntu 20.04, with hardware specifications of an Intel Core i5-13600KF 5.10 GHz processor, 32 GB of RAM, and an NVIDIA GeForce RTX 3090 GPU. The implementation was divided into two stages, with batch sizes set at 32 and 16. The learning rate decayed from \(1\times10^{-3}\) to \(1\times10^{-5}\).We used AdamW\cite{loshchilov2017fixing} as the optimizer, setting \(\alpha\) to 0.9996, the number of heads in the multi-head attention to 8, and the number of points to 2048. The network was trained for 400 epochs during the warm-up phase and 200 epochs during the teacher guidance phase.

\subsection{\textbf{Main Results}}    
\textbf{Baselines.} We compare our method with various baselines and direct extensions of popular 2D image classification semi-supervised methods. First, we conduct qualitative and quantitative experiments on ShapeNet, comparing the proposed model with Pixel2point\cite{afifi2020pixel2point}, 3DAttriFlow \cite{wen20223d}, and AltasNet\cite{groueix2018papier}. Second, we extend state-of-the-art SSL methods for image classification (such as MeanTeacher\cite{tarvainen2017mean}, MixMatch\cite{berthelot2019mixmatch}, and FixMatch\cite{sohn2020fixmatch}) to the 3D reconstruction task as strong semi-supervised baselines. We ensure a fair comparison by using the same backbone and experimental setup for all baselines and our method.

\textbf{Quantitative Comparison Results on Pix3D.} As shown in  \hyperref[table:average_scores]{Table 1}, we present the quantitative comparison results of three supervised learning methods and one semi-supervised learning method using 10\% labeled data on the Pix3D dataset. The supervised methods include Pixel2Point \cite{afifi2020pixel2point}, 3DAttriFlow \cite{wen20223d} and Pix2Voxel\cite{xie2019pix2vox} while our approach (Ours) utilizes a semi-supervised learning strategy.

From the performances evaluated by L1 chamfer distance defined in Eq. (\ref{con:cd}), our semi-supervised method achieved an optimal performance of 6.53, significantly outperforming the other three supervised baseline methods. This superior performance can be attributed to SSMP's multi shape prior fusion strategy, which effectively integrates point cloud information from multiple viewpoints or inputs. This integration enhances the model's understanding of the 3D structure, allowing it to capture geometric features more accurately, thereby significantly reducing reconstruction errors. Specifically, in complex categories such as tool and wardrobe, SSMP demonstrated notably lower Chamfer distances, highlighting its ability to handle intricate shapes effectively.

Additionally, SSMP uses a semi-supervised learning framework that leverages consistency regularization and pseudo-label generation. This enables SSMP to utilize unlabeled data effectively, enhancing the model's generalization capabilities even with limited labeled data. As a result, SSMP consistently achieved lower error values macro to effectively utilize unlabeled data in nearly every category, indicating its more vital generalization ability and robustness for 3D shape reconstruction tasks across different categories. For instance, in the sofa category, SSMP achieved a Chamfer distance of 4.60, compared to 5.68 by 3DAttriFlow, demonstrating its enhanced performance in diverse scenarios.

\begin{table*}
    \centering
    \renewcommand{\arraystretch}{1}
    \caption{3D reconstruction on the Pix3D dataset using 10\% of the data, based on L1 chamfer distance, with the results multiplied by \(10^{2}\) (the lower the better).}
    \label{tab:3d_reconstruction_comparison}

    \scalebox{0.9}{
    \begin{tabular}{c|cccccccc|c}
     \hline 
        \textbf{Method} & \textbf{bed} & \textbf{bookcase} & \textbf{desk} & \textbf{misc} & \textbf{sofa} & \textbf{table} & \textbf{tool} & \textbf{wardrobe} & \textbf{Average} \\
    \hline
        3D attributeflow \cite{wen20223d}& 9.19 & 7.99 & 7.16 & 13.60 & 5.68 & 8.61 & 14.40 & 5.16 & 7.59 \\
        pixel2point\cite{afifi2020pixel2point} & 11.16 & 8.04 & 8.49 & 12.20 & 6.42 & 10.15 & 12.18 & 6.69 & 8.78 \\
        Pix2Voxel\cite{xie2019pix2vox} & 9.47 & 7.43 & 7.87 & 10.21 & 6.37 & 8.49 & 12.39 & 6.04 & 8.53 \\
        \hline
        OURS & 6.73 & 6.48 & 7.11 & 13.44 & 4.60 & 8.07 & 11.66 & 3.42 & 6.53 \\
        \hline
    \end{tabular}}
\end{table*}

\begin{table*}
\centering
    \renewcommand{\arraystretch}{1}
    \caption{3D reconstruction on the ShapeNet dataset using 20\% of the data, based on L1 chamfer distance, with the results multiplied by \(10^{2}\) (the lower the better).}
\label{table:average_scores}
    \scalebox{0.9}{
    \begin{tabular}{c|cccccccccccc|c}
        \hline 
        Method       & Airplane & Bench & Cabinet & Chair & Video & Lamp & Speaker & Rifle & Sofa & Table & Phone & Vessel & Avg \\ \hline
        Pix2Voxel\cite{xie2019pix2vox}  &5.36 & 6.44 & 6.50 & 7.41 & 7.49 & 13.16 & 9.42 & 4.99 & 7.12 & 7.72 & 5.24 & 7.43 & 7.35 \\
        OccNet\cite{mescheder2019occupancy} & 6.74 & 7.14 & 7.02 & 7.90 & 9.67 & 14.46 & 10.33 & 6.48 & 8.22 & 7.51 & 6.02 & 9.78 &8.44\\
        AtlasNet\cite{groueix2018papier} & 5.49& 6.91& 7.26& 7.93& 8.32& 11.52& 9.27& 4.80& 7.86& 6.75& 6.10& 7.51& 7.32\\ 
         Pixel2point\cite{afifi2020pixel2point} & 4.11 & 6.67 & 9.36 & 7.79 & 8.66 & 10.16 & 11.23 & 4.15 & 6.88 & 8.78 & 6.74 & 6.54 & 7.54 \\ 
        3DAttriFlow\cite{wen20223d} & 2.99& 5.37& 7.63& 6.65& 6.55& 7.99& 9.22& 2.96& 5.85& 7.50& 4.82& 4.98& 6.16 \\ 
        \hline    
        SSMP (Ours) & 3.08 & 5.30 & 7.67 & 6.40 & 6.37 & 7.29 & 9.07 & 2.90 & 5.83& 6.85& 4.83 & 4.96 & 5.91 \\ \hline    
    \end{tabular}}
\end{table*}

\textbf{ Qualitative Comparison Results on Pix3D.}
As shown in \hyperref[fig:example4]{Fig. 5}, we present the Qualitative comparison results of three supervised learning methods and one semi-supervised learning method under the setting of 10\% labeled data on the Pix3D dataset. T hese supervised methods include Pixel2Point\cite{afifi2020pixel2point}, Pix2Voxel\cite{xie2019pix2vox} and 3DAttriFlow\cite{wen20223d}.

Overall, our semi-supervised method can generate relatively accurate 3D models. Other methods often produce imprecise shapes in the bed category, whereas our method can accurately reconstruct the correct forms. This advantage primarily stems from our superior feature extraction and data utilization strategies. By employing a semi-supervised learning approach, we effectively leverage a large amount of unlabeled data, enhancing the model's understanding and reconstruction capabilities for the complex structures of beds. Consequently, our method generates more detailed and realistic 3D bed models. Additionally, the application of multi shape prior fusion techniques further improves the ability to capture intricate details of the bed frame and mattress, resulting in models surpassing traditional geometric accuracy and detail representation methods.

In the wardrobe category, other methods typically only outline the general shape of the wardrobe, needing help to restore details such as individual sections accurately. Our semi-supervised approach, however, integrates multi-view point cloud data and employs advanced fusion techniques to reconstruct each detailed wardrobe component. This ensures the generated models closely match the actual objects at the detail level. Similarly, our method demonstrates superior performance in the sofa category by accurately capturing complex details that other methods often overlook. Through in-depth learning of materials and shapes, our model better handles sofas' curves and texture variations, enhancing the realism and intricacy of the generated 3D sofa models.

Although our method achieves lower error rates and overall solid performance in the table and miscellaneous categories, especially when dealing with complex geometric structures, there is a slight increase in error for the miscellaneous category. In the table category, traditional methods often struggle with accurately reconstructing complex geometries. Still, with its enhanced feature learning and multi shape prior fusion strategies, our semi-supervised approach better adapts to the demeanor semi-supervised learning method's enhanced accuracy, generalization, and robustness in reconstructing intricate structures. Despite the marginally higher errors in the miscellaneous category, our model maintains robust overall performance, demonstrating our semi-supervised learning method's enhanced accuracy, generalization, and robustness in 3D shape reconstruction tasks. These results further validate the effectiveness of our approach and its broad applicability across different categories.

\begin{figure*}[ht]  % 使用figure*环境
    \centering
    \includegraphics[width=0.8\textwidth]{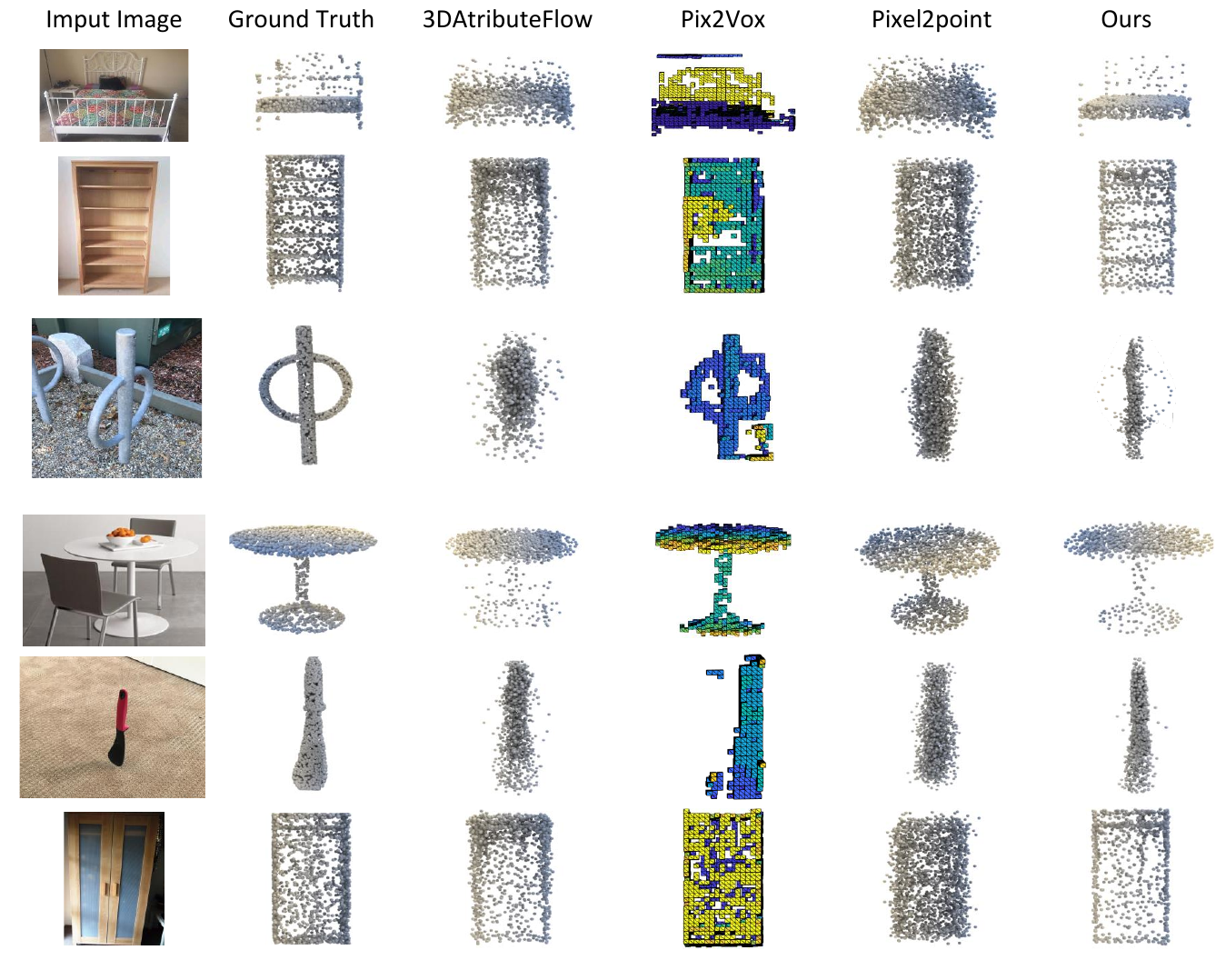}  % 插入PDF文件，宽度为文本宽度的80%
    \caption{Examples of single-view 3D reconstruction on the Pix3D dataset using only 10\% labeled data.}
    \label{fig:example4}
\end{figure*}

\textbf{Quantitative Comparison Results on ShapeNet.} As shown in  \hyperref[table:average_scores]{Table 2}, we present the quantitative comparison results of three supervised learning methods and one semi-supervised learning method using 20\% labeled data on the ShapeNet dataset by L1 chamfer distance described in Eq. (\ref{con:cd}). The supervised methods include Pixel2Point \cite{afifi2020pixel2point}, 3DAttriFlow \cite{wen20223d}, AtlasNet \cite{groueix2018papier}, Pix2Voxel\cite{xie2019pix2vox} and OccNet\cite{mescheder2019occupancy},  while our approach (Ours) utilizes a semi-supervised learning strategy.

Although, from the average results, the performance gain of SSMP shows a specific improvement over some baseline methods (such as 3DAttriFlow), this gain might appear incremental overall. However, a deeper analysis of the specific results across different categories reveals that SSMP consistently achieves lower error values in nearly every category, especially in complex categories (such as Sofa and Vessel) where the error reduction is more significant. This indicates that, despite the seemingly limited overall gain, SSMP demonstrates more robust adaptability and precision in the finer distinctions of different categories.

Furthermore, the semi-supervised learning strategy employed by SSMP has a significant advantage in effectively utilizing unlabeled data. Compared to fully supervised methods that rely entirely on limited labeled data, semi-supervised methods can better capture the inherent structure and distribution of the data, thereby enhancing the model’s generalization ability and robustness. This is particularly important in the ShapeNet dataset, which has high category diversity, as different categories may have significantly different geometric features and complexities.

Lastly, although the performance improvement in specific categories is relatively small, this also reflects the robustness and consistency of SSMP in handling different types of 3D shapes. In contrast, some supervised methods may perform excellently in specific categories but poorly in others, leading to more significant overall performance fluctuations.

\begin{table}[ht]
\centering
    \renewcommand{\arraystretch}{1.5}
    \caption{Using different proportions of data for 3D reconstruction on the ShapeNet dataset, based on L1 chamfer distance, with the results multiplied by 102 (the lower, the better).}
    \label{table:performance_metrics}
    \scalebox{0.9}{
    \begin{tabular}{c|c|c|c}
        \hline
        Method          & \makecell{1\% \\ (361 labels) \\ 36169 images} & \makecell{10\% \\ (3616 labels) \\ 36169 images} & \makecell{20\% \\ (7232 labels) \\ 36169 images} \\ \hline
        Pixel2point\cite{afifi2020pixel2point}    & 9.86              & 7.96                & 7.54\\ 
        3DAttriFlow\cite{wen20223d}    & 7.75              & 6.39                & 6.16\\ \hline
        Ours            & 7.59              & 6.26                & 5.91                \\ \hline
    \end{tabular}}
\end{table}

As shown in \hyperref[table:performance_metrics]{Table 3}, we evaluate the 3D reconstruction performance on the ShapeNet dataset using varying proportions of labeled data (1\%, 10\%, and 20\%). The results demonstrate that increasing the proportion of labeled data consistently reduces the L1 Chamfer distance across all methods. Our semi-supervised approach consistently outperforms others at each data proportion, achieving the lowest L1 Chamfer distance of 5.91 with 20\% labeled data. Notably, even with only 1\% labeled data, our method attains a low error of 7.59, highlighting the effectiveness of the semi-supervised strategy in leveraging limited labeled data for robust learning and generalization.

Our semi-supervised strategy effectively utilizes substantial unlabeled data, enhancing model performance under limited labeled conditions and improving adaptability to diverse shapes and structures. Additionally, our multi shape prior fusion strategy surpasses traditional spherical point clouds by enabling more precise capture of detailed features, resulting in refined and accurate point cloud representations. The integration of these strategies leads to significant reductions in L1 Chamfer distance on the complex and diverse ShapeNet dataset, underscoring the method’s efficiency in data utilization and geometric representation accuracy.

Our method improves as the proportion of labeled data increases, achieving errors of 6.26 and 5.91 at 10\% and 20\% of labeled data, respectively. This performance gain is attributed to the synergistic effects of semi-supervised learning and multi shape prior fusion, fully exploiting the potential of unlabeled data and point feature information. Although the inherent diversity and complexity of the ShapeNet dataset limit the extent of performance improvements, our method demonstrates robust scalability and effectiveness in handling highly complex data, indicating its applicability to a broader range of scenarios.

\textbf{ Qualitative Comparison Results on ShapeNet.} As shown in \hyperref[fig:example5]{Fig. 6}, we present the Qualitative comparison results of four supervised learning methods and one semi-supervised learning method under the setting of 20\% labeled data on the ShapeNet dataset. These supervised methods include 3DAttriFlow\cite{wen20223d},
Pixel2Point\cite{afifi2020pixel2point}, Pix2Voxel\cite{xie2019pix2vox} and OccNet\cite{mescheder2019occupancy}.

Overall, our average shape strategy generates relatively accurate and precise 3D models with fewer noticeable outliers across all categories. Specifically, in the chair category, our method has produced excellent 3D models because the lower regions of angular chairs have been well-learned and constructed. In contrast, other methods struggle to capture the shape of angular chairs accurately. Furthermore, models trained with limited data often exhibit missing parts or shape anomalies, especially in objects such as airplanes and lamps. However, our approach is better at restoring the shapes of these objects.

Additionally, when dealing with complex objects such as airplanes and lighting fixtures, existing supervised methods like OccNet and Pix2Vox often produce 3D models with missing components or anomalous shapes due to limitations in the training data. This issue primarily arises because supervised approaches heavily depend on large volumes of high-quality annotated data. Consequently, these methods struggle to fully capture complex objects' intricate structures and rich details, leading to reduced quality and accuracy in the generated models.

In contrast, our method employs a semi-supervised strategy that integrates a small amount of labeled data with a substantial amount of unlabeled data. This approach enables the model to learn a broader spectrum of shapes and structural patterns. For instance, unlabeled data allows the model to capture detailed features such as wings and engines more accurately, resulting in more complete and realistic 3D models. By leveraging labeled and unlabeled data, our method enhances the model’s ability to generalize across diverse instances, thereby improving the overall fidelity and robustness of the 3D reconstructions.

\begin{figure*}[ht]  % 使用figure*环境
    \centering
    \includegraphics[width=0.7\textwidth]{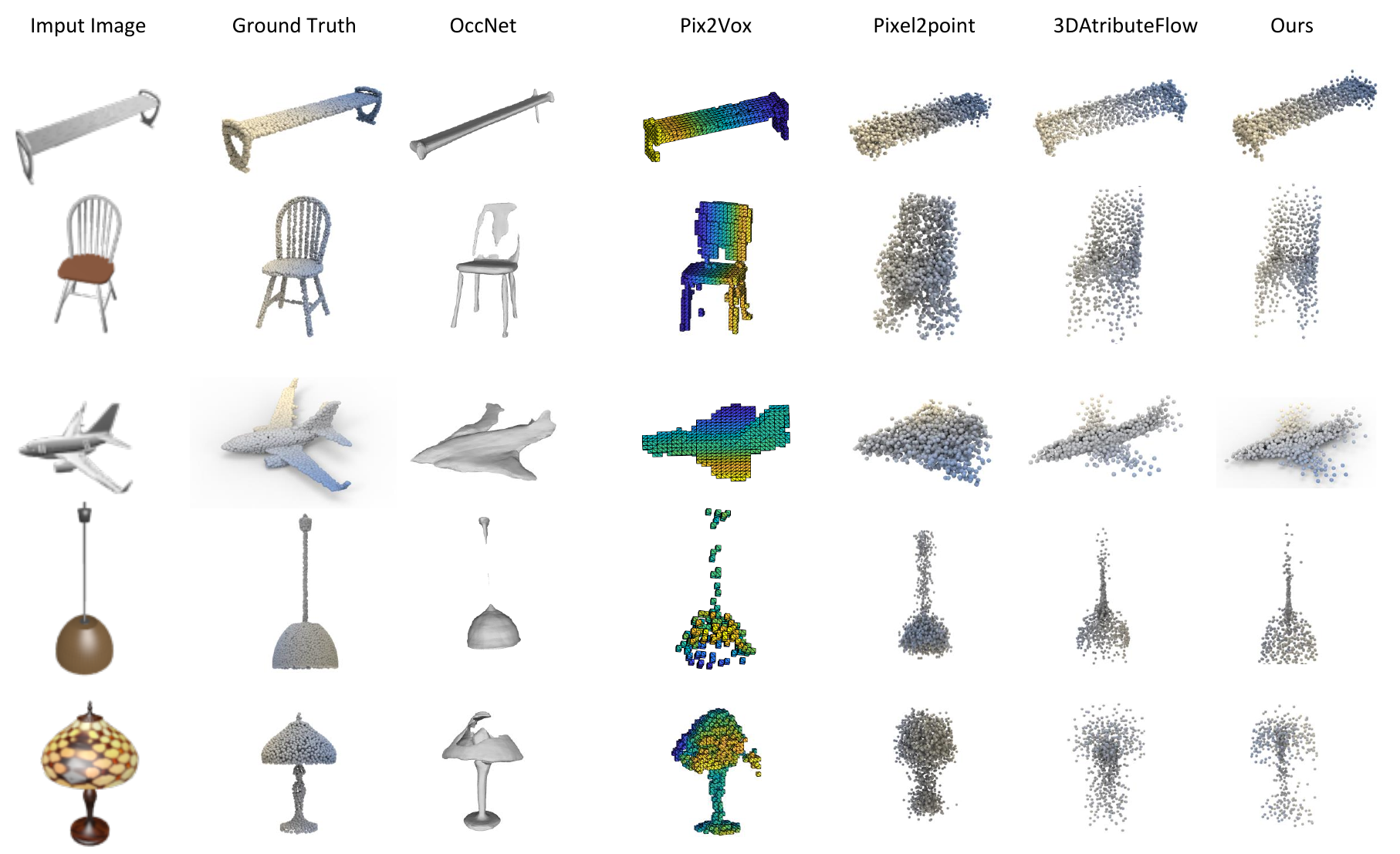}  % 插入PDF文件，宽度为文本宽度的80%
    \caption{Examples of single-view 3D reconstruction on the ShapeNet dataset using only 20\% labeled data.}
    \label{fig:example5}
\end{figure*}

\subsection{\textbf{Ablation Studies}}
In this section, we evaluate the effectiveness of the proposed module and its impact on different point cloud shapes. All experiments were conducted under the 20\% ShapeNet setting unless otherwise noted. 

\textbf{Semi-supervised Strategy.} To analyze the effectiveness of the semi-supervised strategy, we compared our method with several existing semi-supervised methods (including Mean-Teacher\cite{tarvainen2017mean}, MixMatch\cite{berthelot2019mixmatch}, and FixMatch\cite{sohn2020fixmatch}) as well as a model trained for a single stage without using semi-supervised strategies (w/o), as shown in \hyperref[table:cd_results]{Table 4}. Although methods like Mean-Teacher\cite{tarvainen2017mean}, MixMatch\cite{berthelot2019mixmatch}, and FixMatch\cite{sohn2020fixmatch} improved 3D reconstruction performance to some extent, they have limitations in handling the complexity of high-dimensional data, spatial structure, and geometric information. In contrast, our method introduces specific semi-supervised strategies to better leverage unlabeled data, capturing the global structure and details of 3D shapes. It demonstrates significant advantages in 3D reconstruction tasks, achieving an average chamfer distance of 5.91, clearly outperforming other methods. This indicates that semi-supervised learning strategies and model architectures specifically designed for 3D tasks possess greater adaptability and effectiveness in 3D shape reconstruction.
\begin{table}[h]
\centering
    \renewcommand{\arraystretch}{1.5}
    \caption{Comparison of single-view 3D object reconstruction on ShapeNet using different SSL methods. We report the average chamfer distance across all categories.}
    \label{table:cd_results}
        \scalebox{0.9}{
    \begin{tabular}{c|c}
        \hline
        Method Name  & CD   \\ \hline
        Mean-Teacher               & 6.04 \\ 
        MixMatch                    & 6.23 \\ 
        FixMatch                   & 6.10 \\ \hline
        Ours w/o                    & 6.27 \\ 
        Ours                        & 5.91 \\ \hline
    \end{tabular}}
\end{table}

\textbf{Impact of  Multi Shape Prior Fusion Strategy. }To evaluate the effectiveness of the multi-shape prior fusion strategy in 3D shape generation models, we designed a comparative experiment, setting up experimental groups and control groups that used and did not use this strategy, respectively. The experiment used 20\%  of the ShapeNet dataset and 10\% of the Pix3d dataset for the experiment. The control group used only a single shape prior (such as a spherical point cloud) as input. In contrast, the experimental group used a multi-shape prior fusion strategy to generate a fused point cloud as the model input by fusing multiple shape priors. The two models were trained under the same training conditions, and the accuracy of the generated results was measured using the Chamfer distance evaluation metric. As shown in \hyperref[table:model_configurations1]{Table 5}, the network using a single shape prior as input achieved average chamfer distances of 6.03 and 6.71, respectively, while the network using a shape prior fusion strategy achieved slightly better performance with average chamfer distances of 5.96 and 6.53, respectively. This indicates that a single shape prior (such as a spherical point cloud) usually provides limited geometric information and may not be sufficient to describe complex 3D shapes fully. The fusion of multi-shape priors allows the model to receive geometric information from multiple perspectives and different features, forming a richer and more diverse input. This diverse information source helps the model better balance local details and global structure during the generation process, avoiding the limitations of over-reliance on a specific prior

\begin{table}[h]
\centering
    \renewcommand{\arraystretch}{1}
    \caption{Comparison of single-view 3D object reconstruction between single shape priors strategy and multi shape prior fusion strategy on the ShapeNet and Pix3D Dataset. We report the average chamfer distance across all categories.}
    \vspace{0.5em}
    \label{table:model_configurations1}
    \scalebox{0.9}{
    \begin{tabular}{c|cc}
        \hline
        Model Configuration                         & ShapeNet & Pix3D \\ \hline
        Single Shape Priors Strategy    & 6.03  & 6.71\\ \hline
        Multi Shape Prior Fusion Strategy     & 5.91  & 6.53\\ \hline
    \end{tabular}}
\end{table}

\textbf{Impact of Different Multi Shape Prior Fusion  Strategy.} To evaluate the impact of different multi shape prior fusion strategies on model performance, we conducted experiments using three different fusion strategies: 1) weighted averaging using Chamfer distance as the weight, as shown in Eq. (\ref{con:cd1}).

\begin{equation}
 P_o = \frac{\sum_{i=0}^K P_i \cdot d_{CD}(P_i, P_{\text{avg}})}{\sum_{j=0}^K d_{CD}(P_j, P_{\text{avg}})} 
 \label{con:cd1}
\end{equation}

2) simple addition using Chamfer distance as the penalty weight, as shown in \ref{con:cd2}.
\begin{equation}
 P_o = \sum_{i=0}^K  P_i \cdot \frac{1}{d_{CD}(P_i, P_{\text{avg}})} 
 \label{con:cd2}
\end{equation}
As illustrated in \hyperref[table:model_configurations3]{Table 6}, the fusion strategy that employs weighted averaging with Chamfer distance as the weight demonstrates significant effectiveness. Based on the comparison of CD  values in the table, it can be seen that our method (Ours) has a CD value of 5.91, which is better than 6.19 for Strategy F (Eq. (\ref{con:cd1})) and 7.26 for Strategy N (Eq. (\ref{con:cd2})). Because the inverse of the Chamfer distance can better highlight points that are closer to each other, making the fused point cloud more detailed and accurate, while effectively reducing the impact of noise points, thereby improving the reliability and stability of the fusion results.

\begin{table}[h]
\centering
    \renewcommand{\arraystretch}{1.5}
    \caption{Comparison of single-view 3D object reconstruction on ShapeNet using different multi shape prior fusion strategy. We report the average chamfer distance across all categories. F denotes the use of the strategy described in Eq.(\ref{con:cd1}), and N denotes the use of the strategy described in Eq. (\ref{con:cd2}).}
     \vspace{1em}
    \label{table:model_configurations3}
        \scalebox{0.9}{
    \begin{tabular}{c|c}
        \hline
        Method Name  & CD   \\ \hline
        F               & 6.19 \\ 
        N                    & 7.26 \\ 
        \hline
        Ours                        & 5.91 \\ \hline
    \end{tabular}}
\end{table}

\textbf{Impact of Decoder Type.} To verify the effectiveness of the self-attention mechanism decoder, we compared the performance of a standard multi-layer perceptron (MLP) decoder and a self-attention decoder under different model configurations. As shown in \hyperref[table:model_configurations2]{Table 7}, in the ShapeNet and Pix3D datasets, the decoder introducing the self-attention mechanism significantly reduced the average Chamfer distance. It improved the accuracy of 3D object reconstruction. Taking ShapeNet as an example, the Chamfer distance of the Pixel2point model decreased from 7.54 to 7.44, the 3DAttriFlow model decreased from 6.16 to 6.08, and our method further reduced it from 5.98 to 5.91. These improvements demonstrate the advantages of self-attention decoders in capturing long-range dependencies and global features, enabling more effective reconstruction of complex 3D shapes

In the Pix3D dataset, the positive impact of self-attention decoders was also observed. The Chamfer distance of the Pixel2point model decreased from 8.78 to 8.67, the 3DAttriFlow model decreased from 7.59 to 7.45, and our method decreased from 6.62 to 6.53. Although the improvement is relatively small in some cases, the self-attention decoder still achieves a reduction in Chamfer distance. This indicates that the self-attention mechanism is adaptable in different model architectures and can improve 3D reconstruction performance on diverse datasets. Therefore, using self-attention decoders is an effective strategy for optimizing 3D reconstruction results,  enhancing the model's reconstruction ability.

\begin{table}[!h]
\centering
    \renewcommand{\arraystretch}{1}
    \caption{Comparison of single-view 3D object reconstruction on ShapeNet and Pix3D using fusion point cloud with different decoder configurations. We report the average chamfer distance across all categories.}
    \vspace{1em}
    \label{table:model_configurations2}
    \scalebox{0.9}{
    \begin{tabular}{c|ccc}
        \hline
        Model Configuration                     & ShapeNet &Pix3D   \\ \hline
        Pixel2point MLP Decoder    & 7.54  &8.78 \\
        Pixel2point Self-Attention Decoder      &7.44  &8.67 \\
        3DAttriFlow MLP Decoder    & 6.16 &7.59\\  
        3DAttriFlow Self-Attention Decoder      & 6.08 &7.45 \\ \hline
        Ours MLP Decoder    & 5.98 &6.62\\  
        Ours Self-Attention Decoder      & 5.91 &6.53 \\ \hline
    \end{tabular}}
\end{table}
\vspace{-2em}

\section{Conclusion}
We propose the SSMP (Semi-Supervised Multi Shape Prior Fusion Reconstruction) method, which is the first semi-supervised approach for single-view 3D reconstruction of point clouds. To optimize the semi-supervised learning setup, we designed an efficient multi shape prior fusion strategy module to integrate and stabilize feature representations, thereby enhancing reconstruction quality. Additionally, we introduced a self-attention decoder that effectively captures the global features of 3D shapes, resulting in more accurate shape generation. Extensive experimental results on multiple benchmark datasets demonstrate that the SSMP method exhibits outstanding performance in 3D reconstruction tasks. In future work, we plan to further explore the application of other types of 3D representations in semi-supervised learning settings.

\section*{Acknowledgment}

This work is supported by MOE (Ministry of Education in China) Liberal Arts and Social Sciences Foundation 24XJCZH024, General Project of Education Department of Shaanxi Provincial Government under Grant 22JK058, Xi'an Key Laboratory of Aircraft Optical Imaging and Measurement Technology Open Fund Project 2023-006, National Natural Science Foundation of China 62173270 and 52367015, China Postdoctoral Science Foundation under Grant 2024M750897, and Jiangxi Provincial Natural Science Foundation under Grants 20224BAB204051 and 20232BAB214064.

%%Vancouver style references.
\bibliographystyle{cag-num-names}
\bibliography{refs}

\end{document}